\documentclass{article}
\usepackage{spconf,amsmath,graphicx}
\usepackage{todonotes}
\usepackage[ruled,vlined]{algorithm2e}
\usepackage{url}
\usepackage{latexsym}
\usepackage{amssymb}
\usepackage{multirow}
\usepackage{float}
\usepackage{subcaption}
\usepackage{booktabs}
\usepackage{microtype}

\title{A bandit approach to curriculum generation for automatic speech recognition}
\name{Anastasia Kuznetsova$^{\star \dagger}$ \qquad Anurag Kumar$^{\star}$ \qquad Francis M. Tyers$^{\dagger}$}

\address{$^{\star}$ Indiana University, Bloomington\\
            Computer Science Department
        \\
      $^{\dagger}$ Indiana University, Bloomington\\
        Department of Linguistics}
\begin{document}
%
\maketitle
\begin{abstract}
The Automated Speech Recognition (ASR) task has been a challenging domain especially for low data scenarios with few audio examples. This is the main problem in training ASR systems on the data from low-resource or marginalized languages. In this paper we present an approach to mitigate the lack of training data by employing 
Automated Curriculum Learning  in combination with an adversarial bandit approach inspired by Reinforcement learning. The goal of the approach is to optimize the training sequence of mini-batches ranked by the level of difficulty and compare the ASR performance metrics against the random training sequence and discrete curriculum. We test our approach on a truly low-resource language and show that the bandit framework has a good improvement over the baseline transfer-learning model.
\end{abstract}
\section{Introduction}
Automated speech recognition (ASR) is a task of transforming speech signal into text. Lack of training data is problematic in this domain, especially for low-resource languages. By low-resource languages we understand not only endangered languages with fewer native speakers but also the languages which lack digital presence and sufficiently large corpora online. There are efforts towards addressing this problem, creating public-domain speech data for a growing number of languages \cite{Ardila:2019}. However, despite this effort the data problem will almost certainly persist into the foreseeable future as  newly emerging neural architectures are more and more data hungry. Therefore, there is a need for a solution which would mitigate the growing need for data and make speech recognition accessible to speakers of more languages.


The notion of \textit{curriculum learning} is not new in Machine Learning. The first results on the effectiveness of the approach were demonstrated by \cite{Elman:1993}. The idea behind the method is to mimic human study behaviour by presenting to the neural network easy examples in the initial phases of the training and then gradually increasing the difficulty of the examples. \cite{Bengio:2009} conducted the experiments on artificially generated tasks and presented \emph{proof-of-concept} results. Despite the effectiveness of the method the researchers in Natural Language Processing and Speech Recognition have not actively investigated curriculum learning.

Some examples of curriculum learning in speech recognition area include the research by \cite{Suyoun:2017} and \cite{Braun:2017}. While the above studies prove that curriculum learning increases the convergence rate, it also establishes a need for clean speech signals and a higher volume of data. Since we lacked both of these, we propose a method of automated curriculum generation that learns the curriculum online.

\textit{Reinforcement Learning} is an area of Machine learning which exploits three main concepts: agent, environment and reward. The \textit{agent} is acting in the environment and takes decisions based on the \textit{reward} it gets from the environment. The goal of the agent is to maximize the expected reward it can get over the time. Reinforcement learning is widely used in control tasks where the \textit{environment} could be a video game or a real-world visual input as in self-driving cars. 
We define these terms more formally later in Section \ref{sub:problem_def}. 

Kala et al. \cite{Kala:2018} is one of the rare works combining RL and ASR we encountered. The authors train two rival ASR models via the policy gradient method with \textsc{reinforce} style updates.



The remainder of the article is organized as follows: 
Section \ref{sec:methods} formulates the problem in terms of bandit framework and reviews the development of the proposed automated curriculum system; Section \ref{sec:experiments} discusses the experimental setting, training data, evaluation metrics and presents the results of the current research; we reserve Section \ref{sec:discussion} for a brief discussion.

\section{Methodology}
\label{sec:methods}
In reinforcement learning (RL) the idea of curriculum learning is often used in control tasks for games, however it has not been widely used in speech recognition. Our approach was inspired by the work of \cite{Graves:2017} where  \textit{bandit} algorithms are used for automated curriculum generation. 

\subsection{Problem definition}
\label{sub:problem_def}

\subsubsection{A K-armed bandit} A $k$-armed bandit can be formally defined as an agent acting in the reward space $\mathcal{R}$ which aims to collect the maximum expected reward in the finite number of trials. Consider a finite sequence of trials $t = 1, 2, 3 ... T$. 
\begin{enumerate}
\itemsep0pt
    \item During each trial the agent selects the bandit's arm at time $t$, the decision is based on the expected pay-off received previously. We define the arm as action $a_t\in\mathcal{A}_t$ where $\mathcal{A}$ is the set of actions;
    \item After selecting the action $a$ on time step $t$ we observe the reward $r_t\in\mathcal{R}$;
    \item Maximum expected reward defines the value function of each action
  \( q_*(a) =  \mathbb{E}[\mathcal{R}_t|\mathcal{A}_t=a]\)
    which is updated for a played action at every time step after observing the reward\footnote{The asterisk means that $q(a)$ is optimal.}. 
\end{enumerate}


\subsubsection{Curriculum task formulation}

According to the curriculum learning approach we need a way of ranking the examples to define the complexity. \cite{Suyoun:2017} suggest utterance length as the measure of complexity. We decided to take a different approach and ranked the training examples using the compression ratio of the raw audio files.\footnote{We used the standard \texttt{gzip} utility for compressing the distributed WAVs.} The motivation behind is that the noisier the audio file, the more entropy it has, and thus the harder it is to both compress and to learn from. We define compression ratio, which shows by how much (\%) the audio file was compressed compared to its uncompressed counterpart as in (\ref{eq:compression}).
\begin{equation}
    \mathrm{CR} = 1 - \frac{\mathrm{Size}_{\mathrm{after}}}{\mathrm{Size}_{\mathrm{before}}}
    \label{eq:compression}
\end{equation}
\begin{figure}
    \centering
    \includegraphics[width=0.5\textwidth]{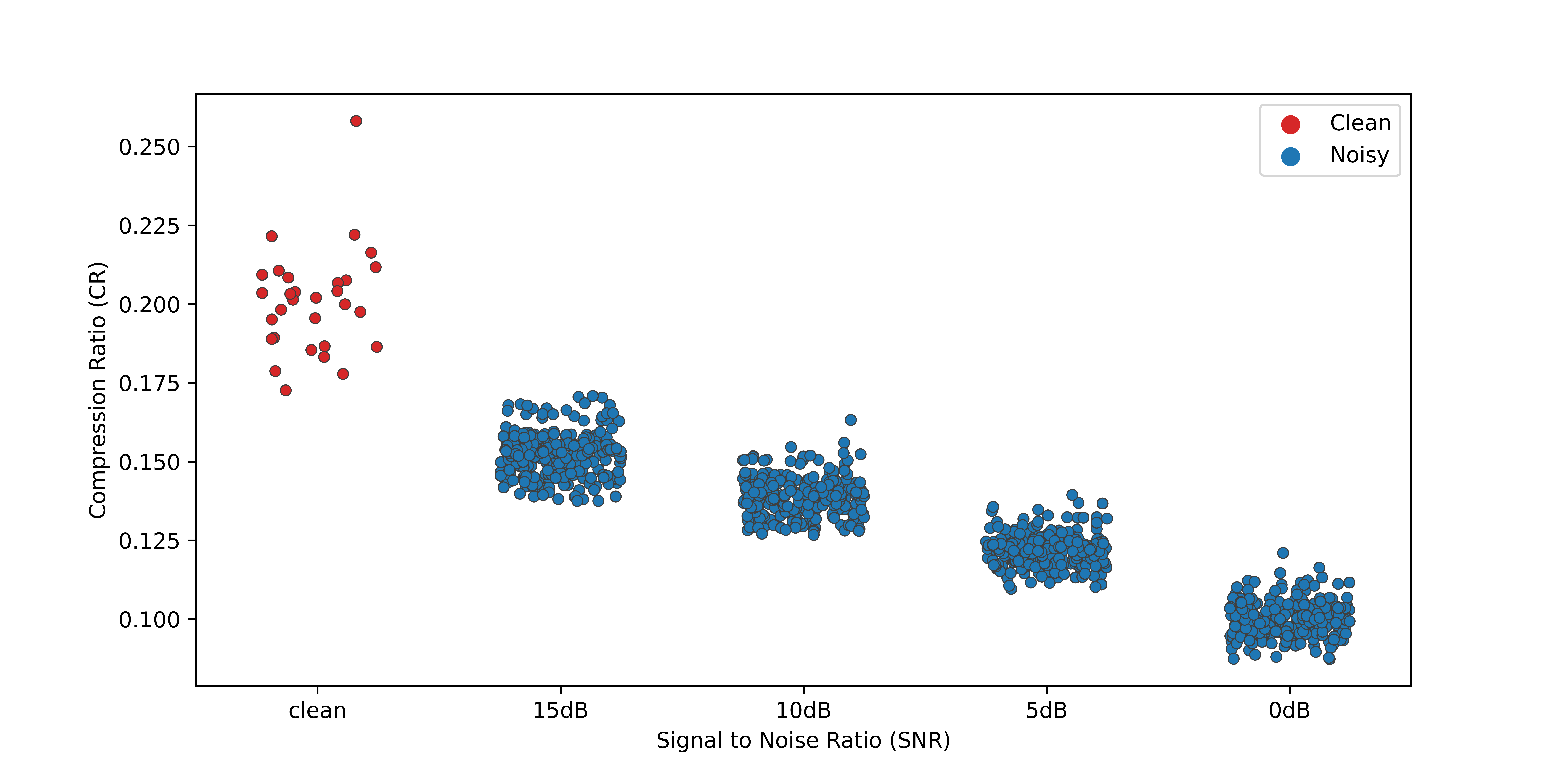}
    \caption{Compression Ratio against different noise levels.}
    \label{fig:snr}
\end{figure}
Figure \ref{fig:snr} demonstrates the dependence of \textit{signal to noise ratio} (SNR) and compression ratio. The plot is based on NOIZEUS corpus \cite{Hu:2007} which contains noisy audio mixtures at different SNRs. The higher the proportion of noise is being blended into speech the less compressed is the audio sample.
Following this logic we divide the data set in $K$ levels of complexity which defines the degree of hardness, where 1 is `easy' and $K$ is `hard'.


Assume the set of input sequences $X$ where every input sequence $x_i\in X$. 

    The \textit{task} is a set of training examples ranked according to the $\mathrm{CR}$. We split the tasks into $K$ categories. Thus, we define a task set as $D = \{D_{1}, D_{2}, D_{3}, ... D_{K}\}$ where $k$ is the index of the task and $x_{i}$ is the training batch sampled randomly from $D_k$.
    The \textit{Curriculum} is a sequence of tasks selected in each training epoch.

The goal is to find the best sequence (curriculum) of batches to maximize the training gain. The action $a\in\mathcal{A}$  (intuitively `pick batch') refers to the task $D_k$ chosen from $D$. This way $a = k$ i.e. the index of the task can be seen as one of the handles of the multi-armed bandit. So at every time step we are going to update the action value function $q(a)$ after receiving the reward. The reward $r(a)$ is calculated as a result of the \textit{progress signals} received from the network. 

\begin{algorithm}
\DontPrintSemicolon
\SetKwInOut{Input}{Input}
\textbf{Initialize:} $w_{i} \leftarrow 1$ (EXP3) or $q(a)$ (UCB1)\;
Create tasks $D_k$\;
\Begin{
\For{$t\rightarrow T$}{
    Draw action index $a$ based on $w_i$ or $q(a)$\;
    $k\leftarrow a$\;
    $batch\leftarrow sample(D_k)$\; 
    Train the model on $batch$\;
    Observe progress gain $\nu$\;
    $r_t \leftarrow Map (\nu)\in [-1,1]$\; 
    Update $w_{i}$ or $q(a)$ on $r_t$
    }
}
\caption{Curriculum Learning}
\label{alg:training}
\end{algorithm}

\subsubsection{Progress gains} Following the authors in \cite{Graves:2017} we seek to experiment with loss-based progress gains such as \textit{prediction gain}: 

\begin{equation}
\label{eq:pg}
    \nu_{PG} = L(x, \theta) - L(x, \theta')
\end{equation}
\noindent
which is the loss before and after training on $k^{th}$ batch. \textit{Self-prediction gain} is defined in Eq. \ref{eq:spg}.

\begin{equation}
     \nu_{SPG} = L(x, \theta) - L(x', \theta') \qquad x'\sim D_k
\label{eq:spg}
\end{equation}
\noindent
SPG assess the gain on the $k^{th}$ training batch and then samples the batch from the same task again to avoid bias. Progress signals are then used as a reward and are re-scaled to fall in the interval $[-1, 1]$ using the mapping in Eq. \ref{eq:scaling}, where $r_t$ is the current reward at time step $t$,  $\hat{r}_t$ is the progress gain at time step $t$, $q_{lo}$ is 0.2 quantile of the gain history and $q_{hi}$ is the 0.8 quantile. 

 In earlier stages of training the difference between the losses is significantly higher than at the end of the training where the magnitude of the losses is not that drastically high due to convergence. The re-scaling is needed to account for this fact  \cite{Graves:2017}. 

\begin{equation}
    r_t = \left\{\begin{array}{lcl}
        -1 & &  \mbox{if} \hat{r}_t < q_{lo} \\
         1 &&  \mbox{if} \hat{r}_t > q_{hi}\\
         \frac{2(\hat{r}_t-q_{lo})}{q_{hi}-q_{lo}} -1 & & \mbox{otherwise}
    \end{array}\right\}
\label{eq:scaling}
\end{equation}

\subsection{Review of bandit algorithms}
In our experiments we use two bandit algorithms: Upper Confidence Bound (UCB1)  \cite{Auer:2002a} and  Exponential-weight Algorithm for Exploration and Exploitation (EXP3) \cite{Auer:2002b}. 

Both of the algorithms demonstrate strong convergence properties and are analytically proven. This fact is of a paramount importance to us since a lot of experiments in Machine Learning hinge on empirically obtained results e.g. hyper parameter tuning and/or are not interpretable.

The bandit algorithms try to solve a well-known exploitation vs. exploration dilemma i.e. how to ensure that the agent selects the best arm most of the time but also explores other options that potentially may lead to a higher expected reward and the least possible regret\footnote{\textit{Regret} is the expected loss which happens because the agent does not always select the best action in favour of exploration.}.

The UCB1 algorithm is based on the notion of \textit{upper confidence index} assigned to each arm \cite{Auer:2002a}. 
Graves et al. \cite{Graves:2017} obtained plausible results using EXP3 so we decided to use it as well to compare its performance against UCB1. Due to space constraints we refer the reader to the definition of the algorithms in the original papers by Aurer et al. \cite{Auer:2002a, Auer:2002b}.


A generalized version of Automated Curriculum Learning incorporating both of the algorithms is shown in Algorithm \ref{alg:training} where $Map$ function is defined as in Eq. \ref{eq:scaling} and the training step is done with the \textit{Baseline 1} model described in Section \ref{subsec:asr_architechture}. $w_i$ is the weight vector required in EXP3 and $q(a)$ represents action-value function in UCB1, $r_t$ is the reward received by the agent at time step $t$.

\subsection{ASR architecture}
\label{subsec:asr_architechture}
The architecture of our ASR system is based on Baidu's DeepSpeech end-to-end speech recognition system \cite{Hannun:2014}, as implemented by Mozilla.\footnote{\url{https://github.com/mozilla/DeepSpeech}} 
The model is a 5-layer Recurrent Neural Network (RNN), where the first three layers have ReLu activation, the fourth layer is a bi-directional RNN and the 5th layer is a non-recurrent  \textit{softmax} output layer predicting the probabilities of the output characters. The network is trained via CTC loss. 

\textit{Baseline 1} system is DeepSpeech with transfer learning on version \texttt{0.5.0-alpha.6} of Mozilla's English model as a source.\footnote{\url{https://github.com/mozilla/DeepSpeech/releases/tag/v0.5.1}}
We slice-off the final 2 layers from the model and resume training for the target language data.
For \textit{Baseline 1} we select batches for training randomly with no curriculum. For the purpose of comparison of the systems with curriculum learning approach the default file-size based sorting in DeepSpeech was turned off. 




\section{Experiments}
\label{sec:experiments}


In this section we describe the experimental setting, algorithm parameters and results. 
We compare the results of the algorithm proposed in Alg.\ref{alg:training} to two \textit{baselines}, the transfer learning model mentioned in Section \ref{subsec:asr_architechture} and the model trained with the conventional method of applying curriculum learning as described in \cite{Suyoun:2017}. \textit{Baseline 2} was trained with transfer learning and is the discrete curriculum model where the data was split in 5 tasks. For both baselines training and development batch sizes are $64$ and $32$, learning rate is $10^{-4}$ and drop out $0.15$.


\subsection{Data set}
We apply Algorthm \ref{alg:training} to Tatar (ISO-639: \texttt{tat}), a Turkic language spoken in central areas of Russia with around 4.3 million first-language speakers \cite{ethnologue:tat}.\footnote{The motivation behind choosing a real low-resourced language instead of artificially creating one using English data is that real datasets better reflect real-world conditions. For example, Tatar has vowel harmony, the orthography is more transparent than English and words are in general longer and fewer. The Tatar-speaking community is interested in speech recognition and we hope that this work will be of direct use to them. } The dataset consists of short utterances from Mozilla's \emph{Common Voice} \cite{Ardila:2019}. For Tatar in the 2019-12-10 release there is 27 hours of speech. The data is split into training, development and test subsets with 7,131, 4,815 and 4,855 examples respectively.\footnote{Note that this dataset split is different from the one in \cite{Ardila:2019}. The precise split from \cite{Ardila:2019} was made on an unpublished version of the corpus which is no-longer available (Meyer, p.c.). Thus the results are unfortunately not directly comparable. The model setup is however the same.} 


\begin{table}[]
\centering
\begin{tabular}{lcrrrr}
\hline
\multicolumn{1}{c}{\textbf{Algorithm}} & \textbf{Gain} &  \textbf{Tasks} & \textbf{Epochs} & \textbf{WER} & \textbf{CER} \\
\hline
Transfer & --- & --- & 25 & 94.62 & 39.75 \\
~ + SWTSK & --- &  5 & 25 & 85.99& 29.85 \\
~ + UCB1 & PG &  5 & 10 & 86.53 & 29.43 \\
~ + UCB1 & SPG &  5 & 10 & \textbf{85.19} & 30.00 \\
~ + EXP3 & PG & 5 & 10 & 85.76 & \textbf{28.84} \\
~ + EXP3 & SPG &  5 & 10 & 85.91 & 30.33 \\
\hline
\end{tabular}
\caption{Experimental configurations and results. The first baseline is transfer learning from English, \emph{Transfer}. The second baseline is \emph{SWTSK}, a switch-task baseline. \textit{PG} stands for prediction gain, \textit{SPG} -- self-prediction gain. The UCB1 algorithm has a parameter of $c = 0.5$ to control the degree of exploration, and EXP3 has a parameter $\gamma = 0.01$ which is the probability of selecting a random action.}
\label{tab:config_results}
\end{table}

\begin{figure}
    \centering
    \includegraphics[width=0.45\textwidth]{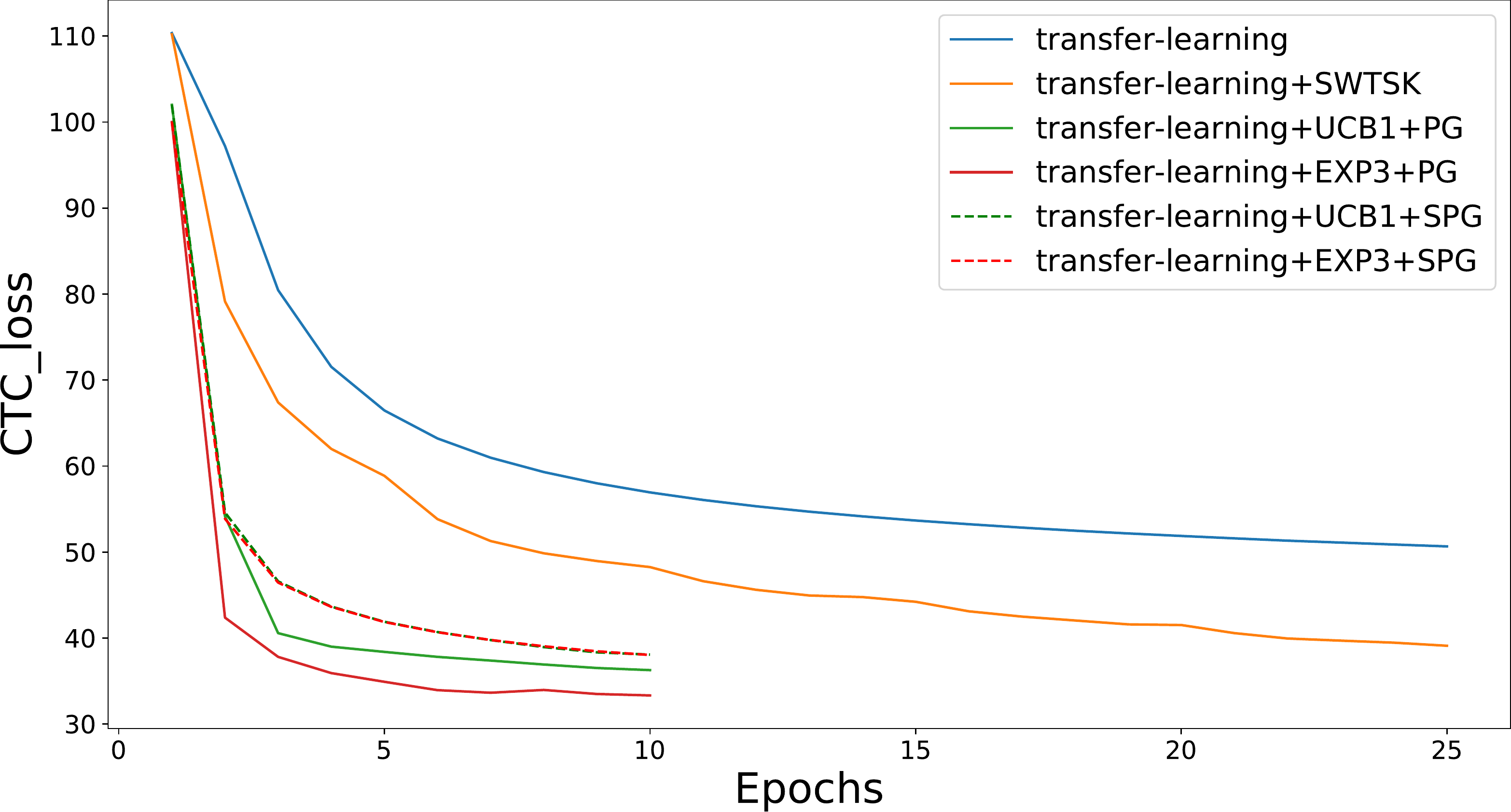}

\caption{Validation loss compared with the baseline models and comparison of cumulative rewards for the bandit models.}\label{fig:val_loss}
\end{figure}
\subsection{Results}

Table \ref{tab:config_results} summarizes the experimental configurations and results for the models. Note that the reported configurations are for the most successful experiments only, although we tested different number of tasks and training for more episodes as well.

The best reduction in WER was achieved by combining UCB1 with \textit{self-prediction} gain while the best CER with EXP3 algorithm and \textit{prediction gain}. The results for WER for both algorithms are comparable i.e. the WER between them does not differ much.

By implementing bandit algorithms we received 10\% reduction in WER and 27\% reduction in CER over \textit{Baseline 1}. Compared to  \textit{Baseline 2} our approach improved WER by approximately 1\% and CER by 3.4\%. The discrepancy is not that big in terms of the evaluation metrics but our models managed to achieve the same result in less time i.e. bandit algorithms contributed to faster convergence leading to better loss reduction. Validation curves on Fig. \ref{fig:val_loss} support this claim: both of the baseline models took 25 training epochs to go on the plateau, bandit algorithms converged only in 10 epochs.

\begin{figure}[ht!]
    \centering
    \includegraphics[width=0.5\textwidth]{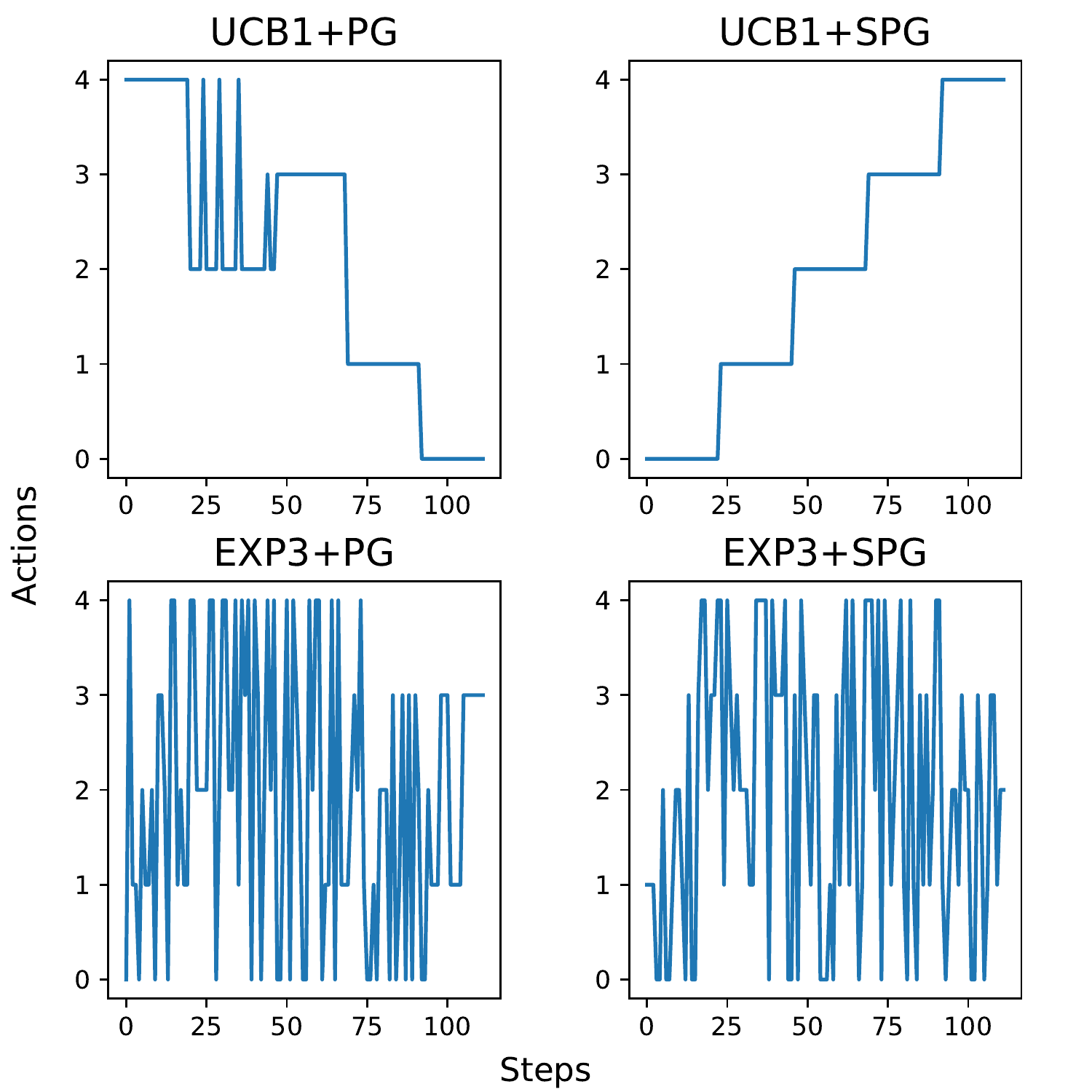}
    \caption{Actions selected by the algorithms in the course of training.}
    \label{fig:actions}
\end{figure}

\section{Discussion}
\label{sec:discussion}

Fig. \ref{fig:actions} demonstrates the action selection during the last $10^{th}$ epoch when the model has reached convergence. Each plot presents the order of the actions taken by the agent. The actions on the \textit{y}-axis show the level of complexity on a scale of $0...4$, where zero is the index of the task with the highest compression ratio and four with the lowest.
The first row in Figure \ref{fig:actions} shows that the action selection is more stable for UCB1 + PG algorithm and the agent follows a certain order of tasks. An interesting observation is that the agent exhausts tasks with lower compression ratios first and tasks with higher CRs later. Whereas UCB1 + SPG actually learns the curriculum that we designed i.e. proceeding through tasks from easy to hard. 
Our hypothesis is that the nature of prediction gains impacts the behaviour of the models.


\textit{Prediction-gain} is a biased estimate of the change in the expected loss on a task (Eq. \ref{eq:pg}). Harder tasks contain longer examples with more noise. We suggest that the model is able to learn better from harder examples especially during early stages of training as these examples contain more information which is captured by PG and thus prefers harder tasks first.

\textit{Self-prediction} gain is an unbiased estimate of the change in the expected loss over a task (Eq. \ref{eq:spg}). When trained on an easier task the model is able to generalize better on the examples sampled from the same simpler task. The reason for that is the higher loss gradient for easier tasks, and thus the model obtains a greater reward in contrast to the harder tasks where the loss gradient, and hence SPG, tends to be lower. 

However, the same trend in the action selection is not observable for the EXP3 algorithm. The generated curriculum is not easily interpretable in terms of the complexity metric we have defined. One possibility is that the batch selection is done following another metric that we have yet to observe in the data. We intend to investigate this as future work.

\section{Concluding remarks}
In this article we have presented several approaches to automatic curriculum development for ASR using bandit algorithms which improve absolute performance in terms of WER and CER and decrease training time. In addition we have outlined a novel measure of training sample complexity --- the compression ratio --- which captures the amount of noise, or entropy. This measure is trivial to apply and substantially improves over random selection of training examples.

\section*{Acknowledgements}

We would like to thank Mansur Saykhunov for his help in creating a language
model for Tatar. We would also like to thank Nils Hjortnæs and Josh Meyer
for their helpful comments.

\bibliography{main}
\bibliographystyle{IEEEbib}

\end{document}